\documentclass[a4paper]{jpconf}
\usepackage[square,sort,compress,numbers]{natbib}
\usepackage{graphicx}
\usepackage{caption}
\usepackage{subfig}
\usepackage{amsmath}

\graphicspath{{figure/}}

\begin{document}
\title{Population-based wind farm monitoring based on a spatial autoregressive approach}

\author{W.\ Lin, K.\ Worden and E.J.\ Cross}

\address{Dynamics Research Group, University of Sheffield, Sheffield S1 3JD, UK}

\ead{wlin17@sheffield.ac.uk}

\begin{abstract}
An important challenge faced by wind farm operators is to reduce operation and maintenance cost. Structural health monitoring provides a means of cost reduction through minimising unnecessary maintenance trips as well as prolonging turbine service life. Population-based structural health monitoring can further reduce the cost of health monitoring systems by implementing one system for multiple structures (i.e.~turbines). At the same time, shared data within a population of structures may improve the predictions of structural behaviour. To monitor turbine performance at a population/farm level, an important initial step is to construct a model that describes the behaviour of all turbines under normal conditions. This paper proposes a population-level model that explicitly captures the spatial and temporal correlations (between turbines) induced by the wake effect. The proposed model is a Gaussian process-based spatial autoregressive model, named here a GP-SPARX model. This approach is developed since (a) it reflects our physical understanding of the wake effect, and (b) it benefits from a stochastic data-based learner. A case study is provided to demonstrate the capability of the GP-SPARX model in capturing spatial and temporal variations as well as its potential applicability in a health monitoring system.
\end{abstract}

\section{Introduction}
In light of the current climate situation and power crisis, 
urgent actions are needed to accelerate the energy transition from fossil fuels to renewables. 
Wind power is a key type of renewable source and currently constitutes 9\% of the world's electricity. 
Although the offshore wind sector has been growing rapidly over the past year, 
the rate of offshore wind installations needs to increase by a factor of four in order to meet the objectives of net zero by 2050 \cite{GWEC2022}. 
The rapid expansion in offshore wind power leads to significant challenges as well as opportunities. 
Given that the cost of operation and maintenance in offshore wind is significantly higher than its onshore counterpart, 
the development of predictive maintenance strategies for wind farms becomes more crucial than ever. 

{\em Population-based structural health monitoring} (PBSHM) provides a means of developing predictive maintenance strategies while accounting for the interactions between the turbines across a wind farm. 
In contrast to the single-structure approach in conventional data-based structural health monitoring (SHM), 
PBSHM focusses on data sharing across a population of structures to improve the predictability of models 
\cite{Worden2015,PBSHM1,PBSHM2,PBSHM3,PBSHM4,PBSHM5,PBSHM6,PBSHM7}. 
When sharing data between structures, one has to acknowledge that no matter how similar the structures are in terms of geometry, material and topology, 
the data collected from different structures are likely to be different because of environmental and operational variations (EOV) and manufacturing tolerances. 
In PBSHM, there are currently two approaches to address this inter-structure difference. 
One of them is based on domain adaptation, 
which removes inter-structure difference by homogenising data from various structures \cite{Gardner2022,Poole2022}. 
The other approach is to model the inter-structure correlations with the help of the physical understanding of EOV \cite{Lin2022dce,PBSHM7}, which is the focus of this study. 

The monitoring of turbine behaviours across an offshore wind farm is a unique problem in many aspects. 
Firstly, the turbines in a wind farm are of the same model, 
meaning that they are expected to behave in the same way under the same environmental and operational conditions. 
Secondly, the turbines are positioned close to each other, sharing the same environment. 
Currently, the most commonly used wind turbines are the horizontal-axis ones, which create a wake region downstream of each turbine rotor. 
The wind environment around each turbine is affected by the existence of wakes, 
therefore, the turbines in a wind farm can affect each other's behaviour through the environment. 
That is, a spatial pattern exists in the wind environment and, thus, turbine behaviours across a wind farm. 
This spatial pattern changes with time-varying wind directions, resulting in spatio-temporal correlations across the farm. 
Capturing these correlations may enable a monitoring method to predict which turbines are more likely to be damaged in a wind farm as well as to maximise the overall power production across the farm. 

In this paper, a two-level method is proposed to model the spatio-temporal correlations across a wind farm. 
The first level consists of several Gaussian process-based spatial autoregressive models (GP-SPARX), 
each capturing the spatial pattern corresponding to a specific wind direction. 
In the second level, (simple) criteria are designed that determine when to switch on/off the first-level models. 
A case study of a simulated wind farm is used to demonstrate the capability of the proposed switching GP-SPARX model, 
which also sheds light on possible ways to improve this method. 

The layout of this paper is as follows. 
Section \ref{sec:two-level} explains the motivation of the proposed two-level method, 
by defining the wind farm modelling problem in detail. 
Section \ref{sec:switching_GPSPARX} provides descriptions for the proposed switching GP-SPARX model. 
Section \ref{sec:case_study} presents the case study and the prediction results obtained using the proposed method, 
with the concluding remarks given in Section \ref{sec:conc}. 

\section{Wind farm modelling as a two-level problem} \label{sec:two-level}
For offshore wind farms with horizontal-axis turbines, the wake effect has a large impact on turbine power production. 
Wake is the region behind a rotor where the wind field is disturbed by the rotating turbine, 
often characterised by decreased mean wind speed and increased turbulence \cite{Archer2018}. 
A downwind turbine placed in the wake of an upwind turbine tends to produce less power than in free-stream conditions. 
The wake-induced power loss can amount to 80\% of the free-stream power in compactly spaced wind farms such as Lillgrund \cite{Dahlberg2009}. 

The existence of wakes gives rise to a spatial pattern across a wind farm. 
As the wind approaches a first-row turbine, the turbine extracts optimal power from the wind, leaving a wake downstream. 
For a wake-shadowed turbine, both its power output and subsequent wake are affected by its upstream wake environment. 
As a result, the turbine behaviours along a path of wake progression are spatially autoregressive. 
The correlation between neighbouring turbines along a wake path is highly nonlinear because of the complex wind-turbine interactions 
-- the wind drives the rotation in a turbine, which then disrupts downstream wind flow. 
To this end, the farm-wise spatial pattern, determined by wake progression paths, is nonlinear and spatial autoregressive by nature. 
A Gaussian process-based spatial autoregressive model is, therefore, suitable to capture this pattern. 

The farm-wise spatial pattern varies with time. 
In a wind farm, although the turbines are programmed to face the incoming wind, they might face slightly different directions due to local turbulence. 
Since the paths of wake progression are highly sensitive to turbine angles, 
a change in any turbine angle will result in a different spatial pattern. 
Therefore, the time-varying turbine angles give rise to time-varying spatial patterns. 
A second-level model is needed to determine how the spatial autoregressive wake progression changes in response to turbine angles. 

In this respect, the modelling of a wind turbine array can be considered a two-level problem: 
the first level concerns the spatial autoregressive wake progression across the farm; 
the second level captures how the spatial autoregressive pattern changes with time-varying turbine angles. 

\section{The switching GP-SPARX model} \label{sec:switching_GPSPARX}

\subsection{GP-SPARX}
Given a fixed set of wake paths, the first level of the model captures the nonlinear, spatial autoregressive variations across a wind farm through a GP-SPARX model \cite{Worden2018gpnarx,Rogers2019thesis,Lin2022ewshm}, 
\begin{align} \label{eq:GPSPARX}
	&u(s,t) = f\Bigl( u_{\infty}(t),\ w(i,s) u(i,t) \Bigr), \quad i=1,\dots,S \\ 
	&f\left( \cdot \right) \sim \mathcal{GP} 
	\left( \mathbf{0}, k_{SE}\left(\cdot,\cdot \right) \right) \nonumber 
\end{align}
where $u(s,t)$ denotes the wind speed at location $s$ and time $t$. 
$u_{\infty}(t)$ is the exogenous input that represents the free-stream wind speed. 
The spatial correlation between turbines is indicated in the weighted autoregressive term $w(i,s) u(i,t)$; 
for every turbine $i$ in the wind farm of $S$ turbines, the weight $w(i,s) = 1$ when the output location $s$ is in the wake of turbine $i$, otherwise $w(i,s) = 0$. 
It ensures that any non-zero autoregressive term corresponds to an upstream neighbour that provides wake shadowing effects to turbine $s$. 
To capture the nonlinearity in wake progression, the function $f\left( \cdot \right)$ is assumed to be a Gaussian process with a mean zero and a squared exponential covariance function. 
For more details on GP regression, the reader is referred to \cite{Rasmussen2006}. 

\subsection{The switching model}
As the turbine angles change, different wake patterns result. 
The idea here is to train a limited number of GP-SPARX models, each corresponding to a spatial pattern, 
and use a higher-level model to switch between these GP-SPARX models when applied to a testing data set. 
In this preliminary study, both the selection of training patterns and the model switching are done manually, 
as a cost-effective way of evaluating the method before developing more complicated/automated alternatives. 
\begin{equation} \label{eq:switching_model}
	u(s,t) =
	\begin{cases}
		f_1\Bigl( u_{\infty}(t),\ w(i,s,\phi) u(i,t) \Bigr) & \underline{\phi}_1 \leq \phi < \bar{\phi}_1 \\
		\qquad \qquad \vdots\\
		f_n\Bigl( u_{\infty}(t),\ w(i,s,\phi) u(i,t) \Bigr) & \underline{\phi}_n \leq \phi < \bar{\phi}_n
	\end{cases}
	,\ i=1,\dots,S 
\end{equation}
In Equation \ref{eq:switching_model}, the output $y$ is obtained by switching between $n$ trained GP-SPARX models $f_1(\cdot),\dots,f_n(\cdot)$, according to criteria based on the wind direction $\phi$. 
Note that both the function $f(\cdot)$ and the weight $w(i,s,\phi)$ change with wind direction. 
$\underline{\phi}$ and $\bar{\phi}$ denote the lower and upper bounds of $\phi$, respectively. 
It indicates that a specific spatial pattern only applies to a certain range of wind directions. 

\section{A case study of a simulated wind farm} \label{sec:case_study}
To evaluate the efficacy of the suggested model form, data from a simulated wind farm are used, 
such that we can assume no inadequate prediction is caused by actual anomalies that may be present in operational data. 

\begin{figure}[ht!]
	\centering
	\subfloat[][Incoming wind at $\pi$ rad]{
		\includegraphics[]{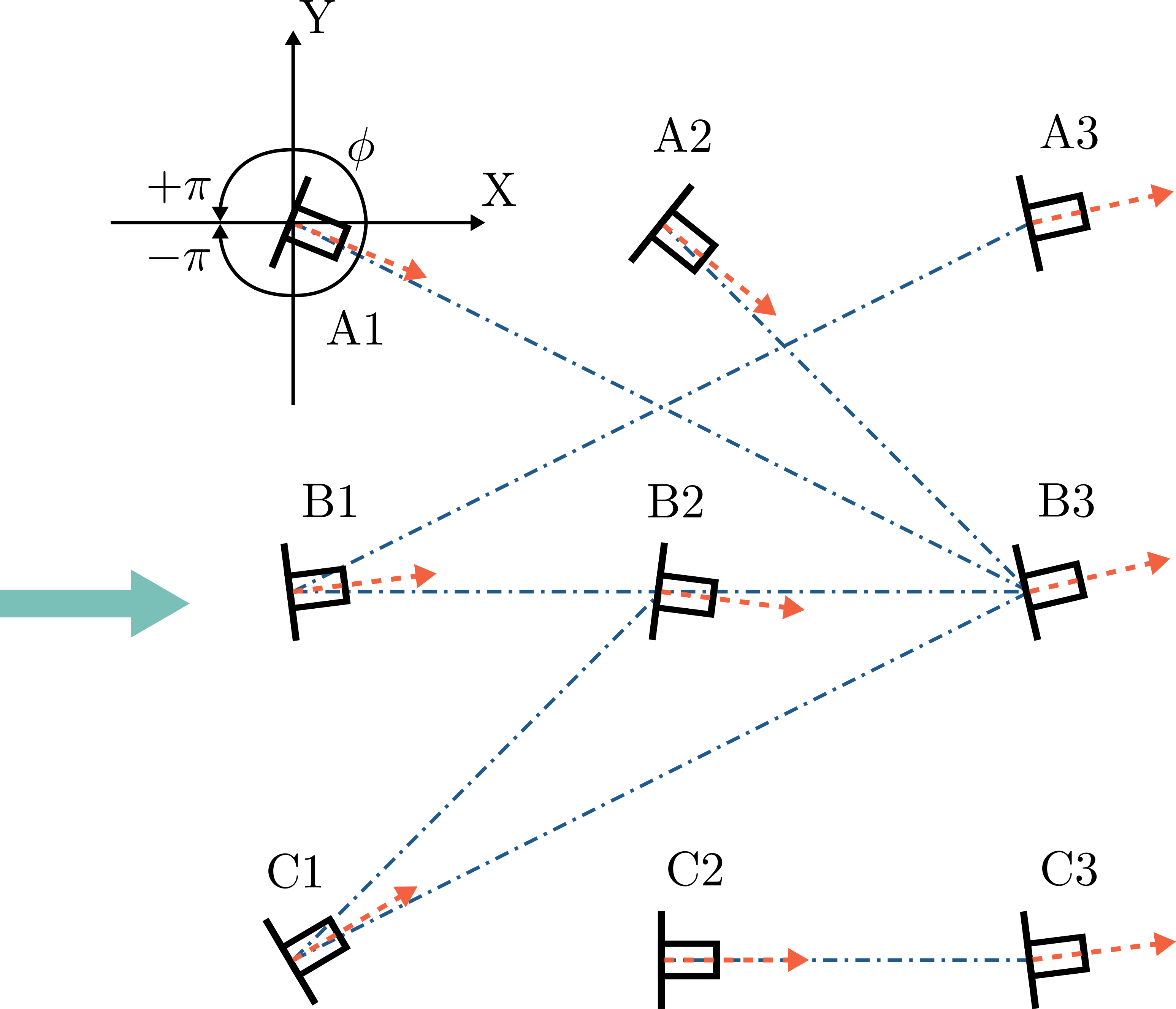}
		\label{fig:sim_farm_270}}
	\hspace{15mm}%
	\subfloat[][Incoming wind at $\pi/2$ rad]{
		\includegraphics[]{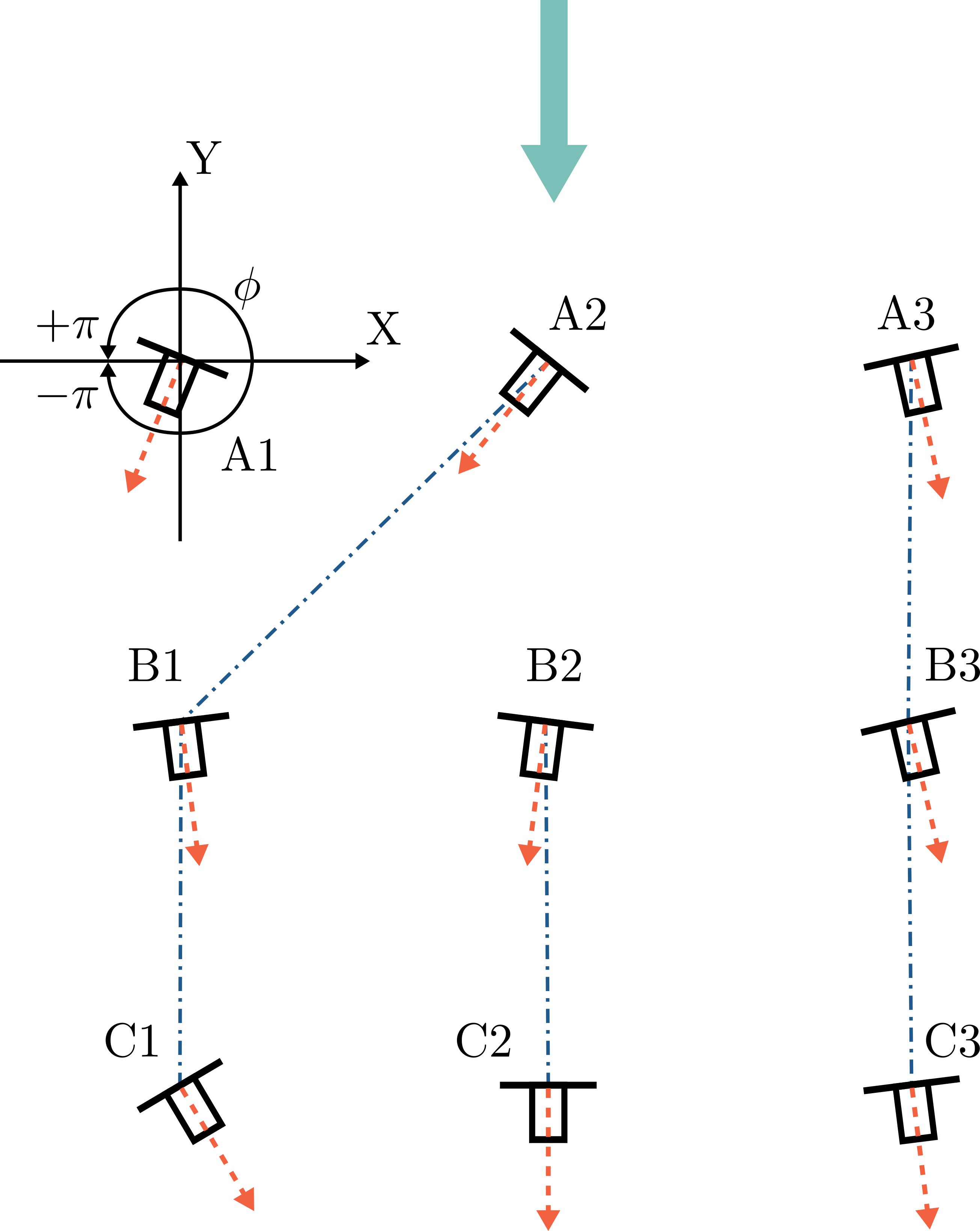}
		\label{fig:sim_farm_0}}
	\caption{Wake patterns in various wind directions.}
	\label{fig:sim_farm}
\end{figure}

The simulated wind farm is shown in Figure \ref{fig:sim_farm}, where the wake patterns associated with two incoming wind directions are illustrated. 
Each blue dash-dotted line indicates a {\em wake path}, along which the wake propagates from an upstream turbine to a downstream turbine. 
The wake path is determined by both the free-stream wind direction and the directions of local air vortices. 

\begin{figure}[ht!]
	\centering
	\subfloat[][Incoming wind at $\pi$ rad]{
		\includegraphics[]{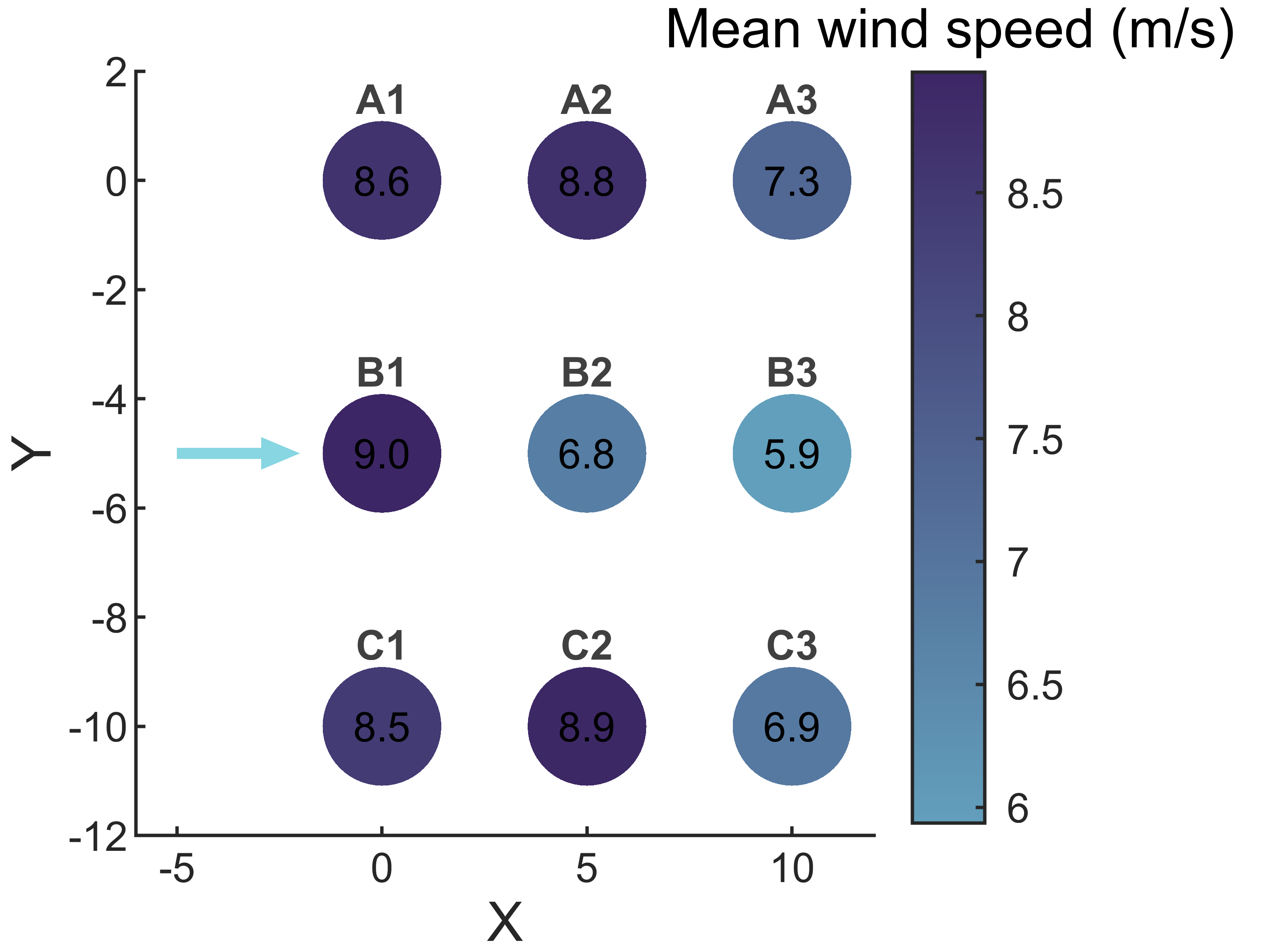}
		\label{fig:mwsp_map_270}}
	\hspace{15mm}%
	\subfloat[][Incoming wind at $\pi/2$ rad]{
		\includegraphics[]{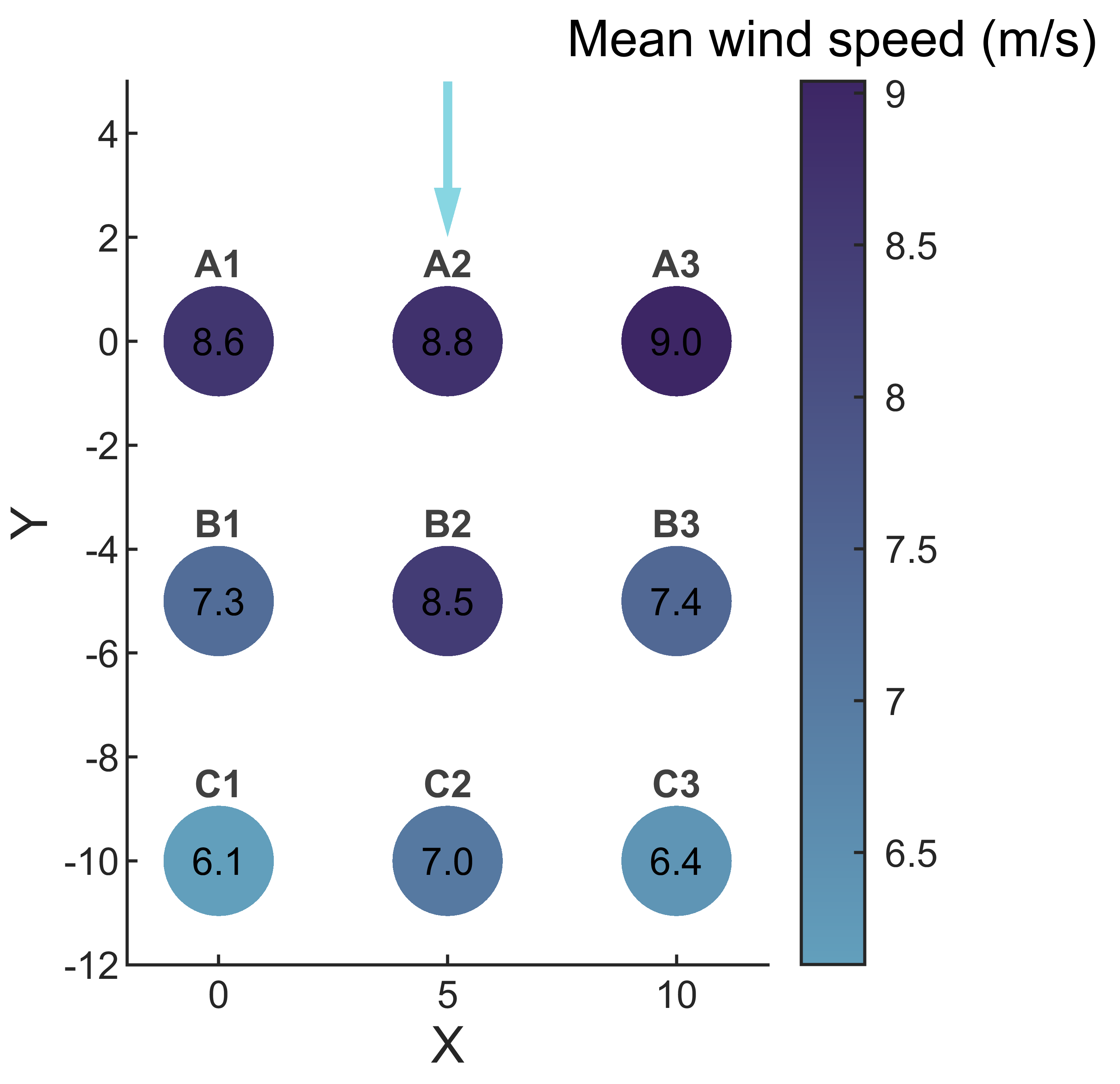}
		\label{fig:mwsp_map_0}}
	\caption{Mean wind speed map in various wind directions.}
	\label{fig:mwsp_map}
\end{figure}

The corresponding spatial patterns of mean wind speed are shown in Figure \ref{fig:mwsp_map}. 
The wind speeds are simulated using a third-order polynomial NARX model \cite{Lin2022ewshm}, 
which provides results that are in line with the most commonly known analytical wake models \cite{Archer2018}. 

\subsection{Training and testing data}
\begin{figure}[ht!]
	\centering
	\includegraphics[]{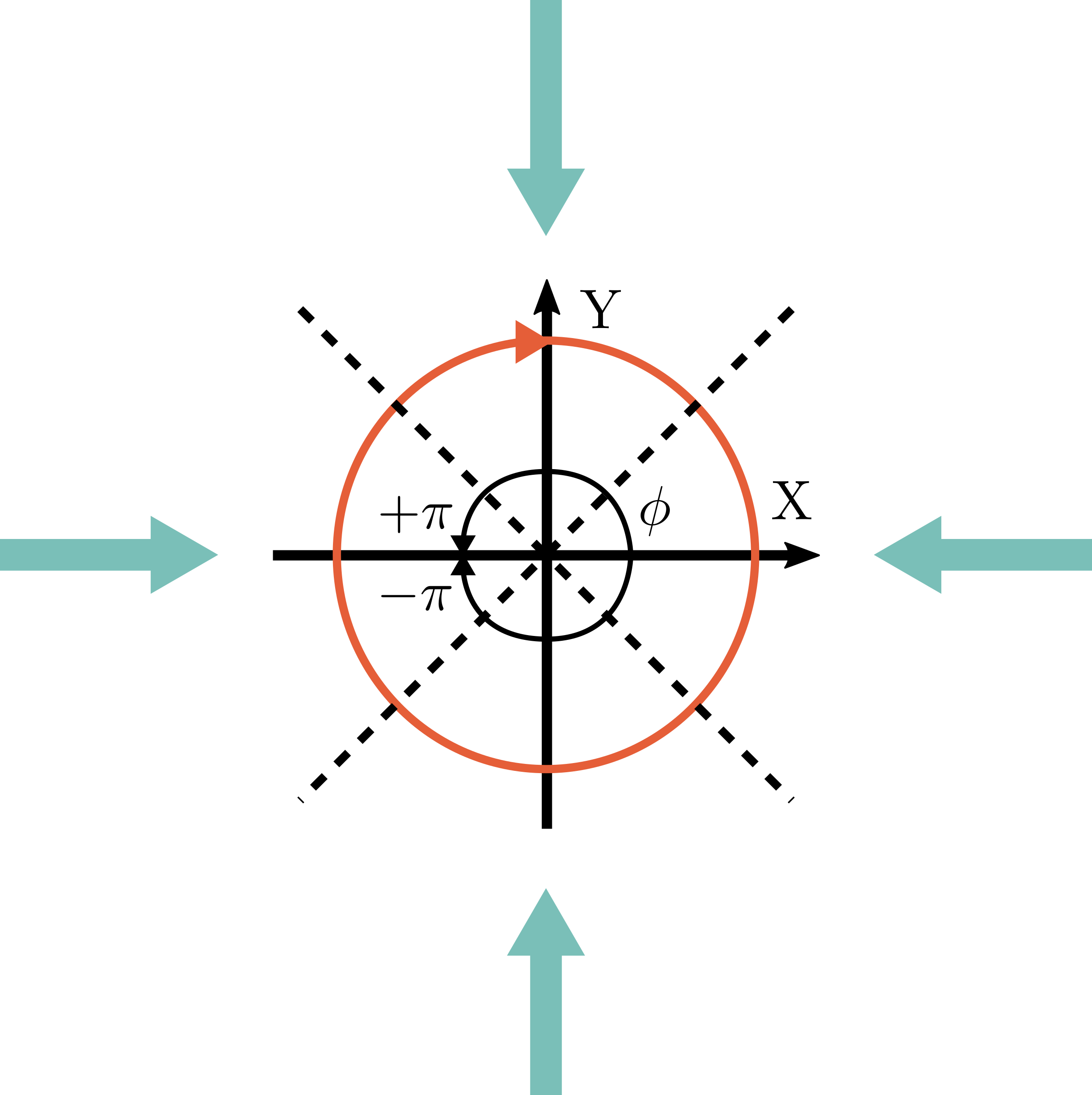}
	\caption{Training and testing data coverage in wind direction.}
	\label{fig:train_test_angles}
\end{figure}

Figure \ref{fig:train_test_angles} demonstrates the training and testing data coverage in terms of wind direction $\phi$. 
In this case study, four wake patterns, associated with four different wind directions (blue arrows in Figure \ref{fig:train_test_angles}), are used to train four first-level GP-SPARX models. 
The testing data, on the contrary, covers the entire range of wind directions, shown as the orange circle arrow in Figure \ref{fig:train_test_angles}. 
Testing predictions are made using the four trained first-level models, by switching between them at predefined boundaries (black dashed lines in Figure \ref{fig:train_test_angles}). 
For simplicity, the boundaries are set as the midpoints between the training angles. 

\subsection{Results}
Previously in \cite{Lin2022ewshm}, the authors have demonstrated that 
the GP-SPARX model is able to capture the nonlinear correlations across a wind farm with satisfactory accuracy 
and outperform deterministic spatial NARX and ARX models in accuracy. 
To carry the study forward, the focus of this paper is to evaluate the efficacy of the second-level switching model in predicting the time-varying spatial patterns across a wind farm. 

\begin{figure}[ht!]
	\centering
	\includegraphics[]{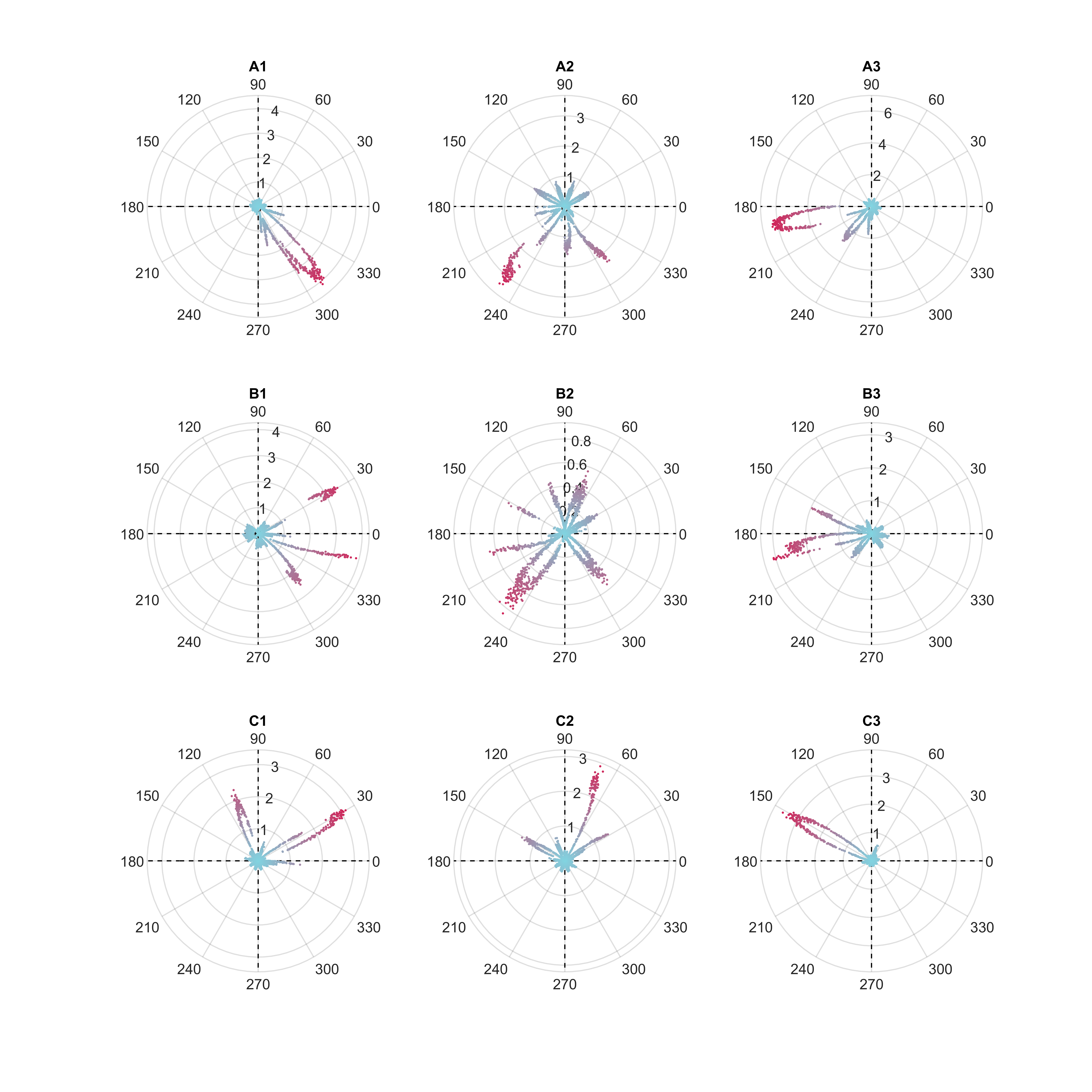}
	\caption{Map of squared errors in polar coordinates.}
	\label{fig:map_polar_err}
\end{figure}

In Figure \ref{fig:map_polar_err}, model prediction errors are presented in polar coordinates to demonstrate the angles at which the two-level model provides satisfactory accuracy or requires improvement. 
Two main observations can be obtained from this error map. 
Firstly, high errors occur in regions away from the training data. 
With the training angles indicated as black dashed lines in Figure \ref{fig:map_polar_err}, 
it is seen that the predictions at training regions are relatively accurate. 
In situations when a considerable amount of error is obtained close to a training region, 
the error reduces rapidly as one approaches the training region. 
Away from the training regions, the model starts to extrapolate (in terms of wind direction), 
thus, there is a tendency for reduced accuracy. 
Secondly, prediction accuracy varies in the regions away from the training data, giving rise to spikes of high errors. 
These spikes indicate the wind directions (i.e.~wake patterns) to be included as part of the training data, to improve the overall accuracy. 

\section{Conclusions} \label{sec:conc}
This paper demonstrates a population-based method that focusses on modelling the correlations between structures, 
for the application of wind farm monitoring. 
One of the main themes of the paper is to encourage the use of physical understanding to simplify modelling approaches. 
For example, the understanding of wake paths provides fixed spatial lags for model training, which circumvents the optimisation problem of spatial lag selection. 
With the rapid growth of computing power, it is crucial not to misspend the computational resources when building data-based models. 
One way to achieve this is to create parsimonious models by exploring physics. 
In the context of wind farm monitoring, the proposed two-level model is potentially able to predict the spatio-temporal correlations across a wind farm, given only one spatial reference (i.e.~weather station measurements). 

This paper provides a preliminary study using an unrefined model form, intending to evaluate the efficacy of the proposed method as well as discovering possible ways to improve the model. 
It is concluded that a switching GP-SPARX model may be able to accurately capture the time-varying spatial patterns in a wind farm if the training regions and switching boundaries are chosen correctly. 
Hence, a possible next step is to use an automatic switching model, such as a treed GP \cite{Worden2013gptree}, 
which learns from data how to select the training regions. 

\section*{Acknowledgements}
The authors would like to acknowledge the support of the EPSRC, particularly through grant reference numbers EP/R004900/1, EP/S001565/1 and EP/R003645/1.
For the purpose of open access, the authors have applied a Creative Commons Attribution (CC BY) licence to any Author Accepted Manuscript version arising. 

\bibliography{ref_mpsva_wl}

\end{document}